\documentclass[10pt,twocolumn,letterpaper]{article}

\usepackage{cvpr}              
\definecolor{cvprblue}{rgb}{0.21,0.49,0.74}
\usepackage[pagebackref,breaklinks,colorlinks,allcolors=cvprblue]{hyperref}

\usepackage{multirow, colortbl, graphicx, multicol, enumitem, float, wrapfig, subcaption}
\usepackage{algorithm, algorithmic, amsmath}

\usepackage[accsupp]{axessibility}

\definecolor{mycolor}{rgb}{0.886, 0.949, 0.996}

\def\paperID{*****} 
\def\confName{CVPR}
\def\confYear{2026}

\title{Cross-Domain Few-Shot Segmentation\\via Multi-view Progressive Adaptation}

\author{Jiahao Nie$^{1,2}$\thanks{Equal contribution} ~\ \ 
Guanqiao Fu$^2$\footnotemark[1] ~\ \ 
Wenbin An$^3$ \ \ 
Yap-Peng Tan$^{4,2}$ \ \ 
Alex C. Kot$^{5,4,2}$ \ \ 
Shijian Lu$^2$\thanks{Corresponding author}\\
$^1$Interdisciplinary Graduate Programme, Nanyang Technological University \\ 
$^2$Nanyang Technological University ~\ \ 
$^3$Xi'an Jiaotong University ~\ \ 
$^4$VinUniversity ~\ \ 
$^5$SMBU\\
{\tt\small jiahao007@e.ntu.edu.sg \ \ shijian.lu@ntu.edu.sg}
}

\begin{document}
\maketitle
\begin{abstract}
Cross-Domain Few-Shot Segmentation aims to segment categories in data-scarce domains conditioned on a few exemplars. Typical methods first establish few-shot capability in a large-scale source domain and then adapt it to target domains. However, due to the limited quantity and diversity of target samples, existing methods still exhibit constrained performance. Moreover, the source-trained model's initially weak few-shot capability in target domains, coupled with substantial domain gaps, severely hinders the effective utilization of target samples and further impedes adaptation. To this end, we propose Multi-view Progressive Adaptation, which progressively adapts few-shot capability to target domains from both data and strategy perspectives. (i) From the data perspective, we introduce Hybrid Progressive Augmentation, which progressively generates more diverse and complex views through cumulative strong augmentations, thereby creating increasingly challenging learning scenarios. (ii) From the strategy perspective, we design Dual-chain Multi-view Prediction, which fully leverages these progressively complex views through sequential and parallel learning paths under extensive supervision. By jointly enforcing prediction consistency across diverse and complex views, MPA achieves both robust and accurate adaptation to target domains. Extensive experiments demonstrate that MPA effectively adapts few-shot capability to target domains, outperforming state-of-the-art methods by a large margin (+7.0\%). Our code is available at \href{https://github.com/niejiahao1998/MPA}{https://github.com/niejiahao1998/MPA}.
\end{abstract}
\section{Introduction}
\label{sec:intro}

Leveraging the paradigm of meta-learning~\cite{chen2019closer,hou2019cross,kang2021relational,kang2019few,xiao2022few,zhang2022meta}, Few-Shot Segmentation (FSS) has achieved great success by learning and adapting few-shot capability from base categories to novel categories~\cite{snell2017prototypical,dong2018few,tian2020prior,fan2022self,min2021hypercorrelation,peng2023hierarchical,xiao2024cat}. FSS typically requires abundant training samples from base categories to establish the capability for segmenting query images conditioned on a few provided support images. However, acquiring abundant base-category samples is often challenging in various data-scarce domains such as medical images~\cite{codella2019skin,candemir2013lung} and satellite images~\cite{demir2018deepglobe}. Enabling FSS in data-scarce domains is a critical and valuable endeavor, driven by its broad range of real-world applications.

\begin{figure}[t]
    \centering
    \vspace{-6mm}
    \includegraphics[width=\linewidth]{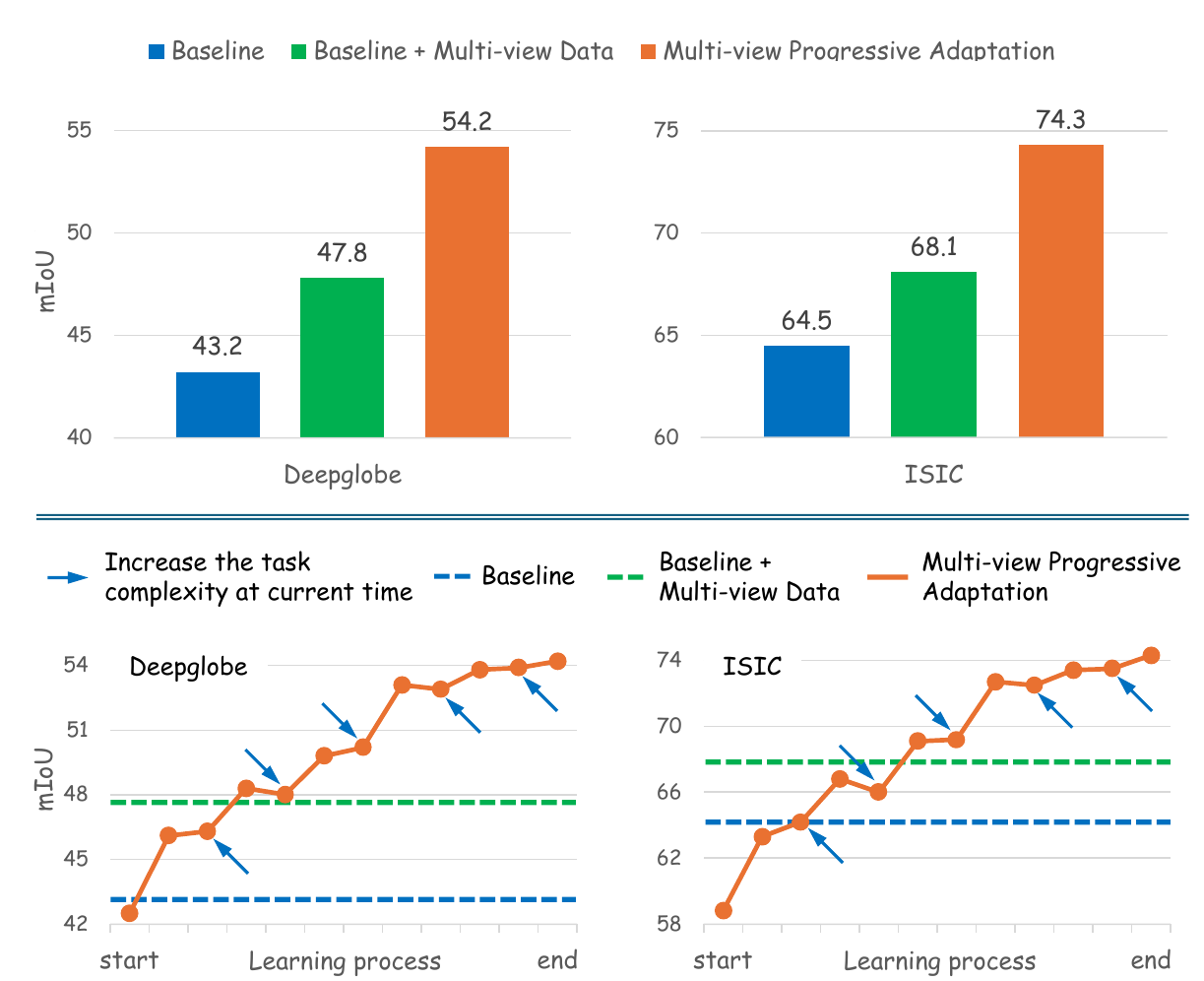}
    \vspace{-5mm}
    \caption{\textbf{Up:} Simply incorporating multiple augmented views from the accessible target samples increases the sample available for establishing few-shot capability but yields only marginal gains, as the large domain gap limits effective utilization of heavily perturbed views. In contrast, our proposed Multi-View Progressive Adaptation (MPA) significantly improves the performance. \textbf{Down:} MPA adopts a progressive strategy to address the challenges posed by large domain gaps. Specifically, it starts with an easy task and progressively increases task complexity as the model becomes more capable during adaptation. This design enables a smooth adaptation of source-trained model and effectively establishes few-shot capability in the target domains.}
    \vspace{-3mm}
    \label{fig:motivation}
\end{figure}

Cross-Domain Few-Shot Segmentation (CD-FSS)~\cite{lei2022cross,he2024apseg,su2024domain,tong2024lightweight,nie2024cross} has been explored to tackle this challenge by leveraging a large-scale source domain to obtain the abundant training samples of base categories. CD-FSS methods typically adopt two learning stages: \textit{(i)} meta-training over a large-scale source domain (such as Pascal VOC~\cite{everingham2010pascal}) with abundant base-category samples for establishing few-shot capability; and \textit{(ii)} adapting this capability to each data-scarce target domain with limited accessible exemplars. However, this paradigm faces two critical challenges: \textit{(i)} only extremely limited samples in target domains is available for adaptation; and \textit{(ii)} substantial domain gaps between the source and target domains hinder the effective adaptation of the few-shot capability. Consequently, the baseline method~\cite{lei2022cross} yields limited performance in target domains, as shown in Fig.~\ref{fig:motivation}~(Up).

Motivated by the observation that CD-FSS methods typically achieve higher performance in multi-shot settings than in the 1-shot setup, we attempt to tackle the aforementioned challenges by augmenting more views from the accessible samples for adaptation. As illustrated in Fig.~\ref{fig:motivation}~(Up), incorporating multiple views yields performance gains but still remains suboptimal. This limitation stems from the large domain gaps, which hinder the establishment of few-shot capability under heavily perturbed augmentations during the early adaptation stage, when the source-trained model exhibits weak few-shot capability in target domains. Consequently, the effective utilization of augmented views is restricted, leading to limited performance. These observations highlight the necessity of designing an appropriate adaptation strategy to fully exploit augmented views.

Inspired by the success of progressive strategies in domain generalization~\cite{zhang2019curriculum,lian2019constructing,sakaridis2019guided,wang2021survey}, we propose Multi-view Progressive Adaptation (MPA) to gradually tackle CD-FSS challenges. Specifically, MPA progressively increases task complexity and diversity to establish few-shot capability from both the data and strategy perspectives. \textbf{\textit{First}}, from the data perspective, we design Hybrid Progressive Augmentation (HPA) that progressively increases the complexity and diversity of the augmented views as adaptation proceeds and the model becomes more capable: \textit{(i)} generating more challenging query views through the accumulation of multiple strong augmentation strategies; \textit{(ii)} increasing the number of augmented views to establish support–query correspondences under more diverse and challenging situations. \textbf{\textit{Second}}, from the strategy perspective, we develop Dual-chain Multi-view Prediction (DMP) to effectively exploit the progressively complex augmented views, which learns through two complementary prediction chains: \textit{(i)} a sequential chain, which progressively extends support–query correspondences~\cite{nie2024cross} across all augmented views, where prediction errors are propagated and accumulated in later views; \textit{(ii)} a parallel chain, which performs multiple ``support-to-query'' predictions across the augmented views, where diverse predictions lead to varied errors. DMP applies supervision to regularize these accumulated and diverse errors, thereby facilitating the establishment of few-shot capability. Thanks to these designs, MPA progressively brings performance gains with each increment in task complexity and successfully establishes few-shot capability in target domains by the end of adaptation (Fig.~\ref{fig:motivation}~(Down)). Extensive experiments along with in-depth analyses demonstrate the superiority of MPA, which effectively addresses the challenges of domain gaps and limited data accessibility in CD-FSS.

The contributions of this work can be summarized in three aspects: \textit{\textbf{First}}, We identify two key constraints in CD-FSS task: (i) the accessible target samples for adaptation is limited in both quantity and diversity; and (ii) the source-trained model exhibits weak few-shot capability in target domains due to substantial large domain gaps. \textit{\textbf{Second}}, we propose Multi-view Progressive Adaptation (MPA), which progressively augments views with increasing complexity and diversity from accessible samples and establishes few-shot capability in target domains from both data and strategy perspectives. \textit{\textbf{Third}}, extensive experiments demonstrate that MPA achieves strong performances over state-of-the-art methods (+7.0\%) and enhances learning efficiency.
\section{Related Work}
\label{sec:related}
\noindent\textbf{Few-Shot Segmentation (FSS)}~\cite{shaban2017one,zhang2021few,cheng2022holistic,luo2023pfenet++,zhu2024addressing,moon2023msi,xu2023self,yang2023mianet,xu2024hybrid,xu2024eliminating,qiu2024aligndiff}, a topic that has been extensively studied, focuses on segmenting novel categories using only a few support images. Existing methods can generally be categorized into two types. Prototype-based methods~\cite{snell2017prototypical,dong2018few,wang2019panet,tian2020prior,li2021adaptive,liu2022intermediate,lang2022learning,lang2023base,fan2022self} segment masks by computing similarities between all query features and support prototypes. In contrast, affinity-based methods~\cite{lu2021simpler,min2021hypercorrelation,zhang2021few,peng2023hierarchical} establish dense correspondence between query and support, heavily relying on rich contextual information. While these approaches are well-established, their robustness under cross-domain setups remains suboptimal~\cite{wang2022adaptive,nie2024cross}.

\noindent\textbf{Cross-Domain Few-Shot Segmentation (CD-FSS)} tackles the FSS setup applied in data-scarce target domains~\cite{lei2022cross, lu2021simpler,boudiaf2021few,wang2022remember,tong2025self,fu2024cross,wu2024task,chen2024cross,chen2024pixel}, aiming to adapt models trained on a large-scale source domain to diverse data-scarce target domains. CD-FSS presents significant challenges due to the following factors: \textit{(i)} during the adaptation stage, the limited target samples increases the risk of overfitting; and \textit{(ii)} the source and target domains exhibit substantial domain gaps. Early works explore this challenge through approaches from dynamic adaptation refinement~\cite{fan2023darnet} and knowledge-transfer~\cite{huang2023restnet} aspects, but these methods exhibit limited generalization capabilities. 
Recently, efforts have been made to adapt source-trained models by disentangling the feature frequency~\cite{tong2024lightweight} and by fine-tuning specific structures rather than the entire model to improve the robustness and mitigate overfitting~\cite{fan2024adapting,su2024domain}. Furthermore, some recent works~\cite{zhang2023personalize,yang2024tavp,he2024apseg,peng2025sam,nie2026boosting} leverage the strong generalization capability of SAM~\cite{kirillov2023segment}, demonstrating improved performance. However, ABCDFSS~\cite{herzog2024adapt} argues that training on the source domain may hinder the development of few-shot capability in the target domain by introducing additional domain gaps. Therefore, this work places greater emphasis on the adaptation stage and explores how to leverage the limited accessible samples to adapt the source-trained model and establish few-shot capability in target domains.

\section{Preliminary Study}\label{sec:preliminary_study}

\subsection{Problem Formulation}
Few-Shot Segmentation aims to segment novel categories using category-agnostic knowledge learned from abundant base-category samples. However, obtaining sufficient training samples is infeasible in data-scarce domains, leading to a two-stage Cross-Domain Few-Shot Segmentation (CD-FSS) paradigm: It first meta-trains the backbone model with a large-scale source domain \boldmath\(\mathcal{D}_{\text{source}}\) to establish few-shot capability, and then adapts the few-shot capability to each data-scarce target domain \(\mathcal{D}_{\text{target}}\) separately. More details are in the supplementary materials.

Following the meta-learning framework~\cite{lei2022cross}, we utilize an episodic training strategy. For the \unboldmath$K$-shot setting, each episode consists of a support set \(S = \{(I_s^i, M_s^i)\}_{i=1}^K\) and a query set \(Q = \{(I_q, M_q)\}\), containing samples from the same category. Here, \(I_s\) and \(I_q\) represent support and query images, while \(M_s\) and \(M_q\) denote their ground-truth masks. The framework consists of two stages: \textit{(i)} training the backbone model in \boldmath$\mathcal{D}_{\text{source}}$ with both \unboldmath$S$ and $Q$ sets; and \textit{(ii)} fine-tuning the trained model to \boldmath$\mathcal{D}_{\text{target}}$ with \unboldmath$S$ only.

\begin{table}[t]
    \renewcommand\arraystretch{1.1}
    \centering
    \vspace{-6mm}
    \caption{The mIoU (\%) of views with more complex augmentation operations is consistently lower than views with simpler augmentations, indicating that the task becomes more difficult.}
    \vspace{-2mm}
    \resizebox{\linewidth}{!}{
    \setlength{\tabcolsep}{8pt}
    \begin{tabular}{l|c|c}
        \toprule[1pt]
        Augmentation& Deepglobe& ISIC\\\hline
        Flip& \cellcolor{mycolor}\textbf{53.1}\ \ & \cellcolor{mycolor}\textbf{71.1}\ \ \\
        Flip+hue variation& 47.9$_{\downarrow}$\ & 69.8$_{\downarrow}$\ \\
        Flip+hue variation+brightness change&  44.5$_{\downarrow\downarrow}$&64.4$_{\downarrow\downarrow}$\\
        \bottomrule[1pt]
    \end{tabular}
    }
    \label{tab:hca_aug}
    \vspace{-3mm}

\end{table}

\subsection{Comparison of Difficulty across Multiple Views}
We increase the task difficulty through two designs: stronger data augmentations and sequential predictions across multiple views. The verification is provided below.

\noindent \textbf{Difficulty definition.} We define difficulty from a quantitative perspective. Specifically, we perform ``support-to-query'' inference with different query images, where lower segmentation performance indicates higher task difficulty.

\noindent \textbf{Stronger data augmentations.} We hypothesize that cumulatively applying more complex augmentations introduces stronger perturbations, thereby increasing the difficulty of predicting accurate masks on the augmented views. For example, an easy view involves only a horizontal flip, a harder view combines flipping with hue variation, and adding brightness adjustment further increases the difficulty. We experimentally validate this assumption in Tab.~\ref{tab:hca_aug}. 

\noindent\textbf{Sequential prediction across multiple views.} We assume that errors accumulate and propagate across sequential predictions. Specifically, the prediction for the first query is guided by the support image, while subsequent queries are conditioned on the predictions of their preceding ones. This assumption is experimentally validated in Tab.~\ref{tab:dmp_seq}, showing that sequentially leveraging multiple augmented views provides a promising way to increase task difficulty.

\begin{table}[t]
    \renewcommand\arraystretch{1.1}
    \centering
    \vspace{-6mm}
    \caption{The mIoU (\%) of later views in the sequential prediction chain is consistently lower than that of earlier views, indicating that errors are propagated and accumulated.}
    \vspace{-2mm}
    \resizebox{\linewidth}{!}{
    \setlength{\tabcolsep}{15pt}
    \begin{tabular}{l|c|c|c}
        \toprule[1pt]
        Dataset& $1^{st}$ view& $3^{rd}$ view& $6^{th}$ view\\\hline
        Deepglobe& \cellcolor{mycolor}\textbf{53.1}& 52.0$_{\downarrow}$& 49.6$_{\downarrow\downarrow}$\\
        ISIC& \cellcolor{mycolor}\textbf{71.1}& 70.1$_{\downarrow}$& 68.5$_{\downarrow\downarrow}$\\
        \bottomrule[1pt]
    \end{tabular}
    }
    \label{tab:dmp_seq}
    \vspace{-3mm}

\end{table}

\begin{figure*}[t]
    \centering
    \vspace{-6mm}
    \includegraphics[width=0.97\linewidth]{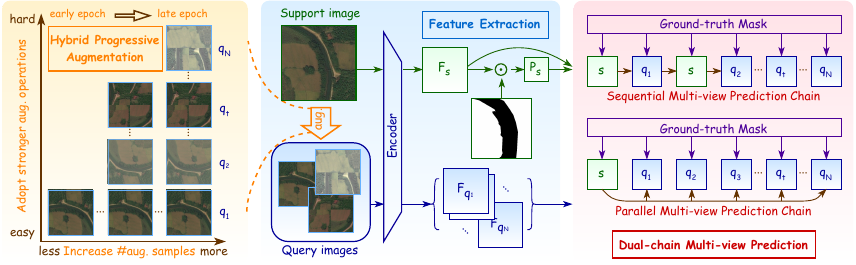}
    \vspace{-2mm}
    \caption{The framework of the proposed Multi-view Progressive Adaptation (MPA). MPA starts with Hybrid Progressive Augmentation (HPA) as highlighted in the yellow box, which introduces two strategies to progressively increase the complexity and diversity of augmented query images. Leveraging the augmented views, MPA establishes few-shot capabilities in target domains via Dual-chain Multi-view Prediction (DMP), incorporating sequential and parallel chains as highlighted in the red box. Note $s$ and $q_i$ denote support and $i^{th}$ query images, $F_s$ and $F_{q_i}$ denote the support and the $i^{th}$ query features as extracted by the encoder, and $P_s$ denotes the support prototype.}
    \vspace{-3mm}
    \label{fig:framework}
\end{figure*}

\section{Method}
\label{sec:method}

To adapt the source-trained model and progressively establish few-shot capability in data-scarce target domains~\cite{wang2021survey}, we propose Multi-view Progressive Adaptation using multiple augmented views from accessible samples~\cite{samarasinghe2023cdfsl} with well-designed strategies, as illustrated in Fig.~\ref{fig:framework}. The proposed framework consists of two major designs: \textit{(i)} Hybrid Progressive Augmentation (HPA): progressively augment more views with increasing complexity and diversity as the adaptation proceeds; and \textit{(ii)} Dual-chain Multi-view Prediction (DMP): establish few-shot capability through both sequential and parallel chains with augmented views. The basic pipeline can be formulated as follows: Given an input support image and its corresponding mask \unboldmath$\{(I_{s}, M_{s})\}$, we generate $N$ augmented query images with masks, denoted as \(\{(I_{q_i}, M_{q_i})\}_{i=1}^N\), through adaptive augmentation operations upon progressive criteria. Next, all support and query images are processed through a weight-shared encoder to extract features \(\{F_{s}, \{F_{q_i}\}_{i=1}^N\}\). Then, the support feature \(F_s\) and its mask \(M_s\) are processed using masked average pooling to generate the support prototype \(P_s\). Finally, \(P_s\), \(F_s\), and \(\{F_{q_i}\}_{i=1}^N\) are utilized to progressively establish few-shot capability in data-scarce target domains. We elaborate our designs as follows: HPA and DMP are detailed in Sec.~\ref{ssec:hca} and Sec.~\ref{ssec:dmp}, respectively. Furthermore, the loss functions are explained in Sec.~\ref{ssec:loss}. Finally, extension to $K$-shot setting is in Sec.~\ref{ssec:kshot}.

\subsection{Hybrid Progressive Augmentation}\label{ssec:hca}

In the target domains, only one support image $I_s$ and corresponding ground-truth mask $M_s$ are accessible under the 1-shot setting, thus $N$ query images \(\{I_{q_i}\}_{i=1}^N\) with corresponding labels \(\{M_{q_i}\}_{i=1}^N\) are derived from support image via augmentation operations:
\vspace{-1mm}
\begin{equation}
    \{(I_{q_i},M_{q_i})\}_{i=1}^N = \mathcal{AUG}(I_{s}),\ \mathcal{AUG}(M_{s}).
\end{equation}

The progressive strategy refers to ``learning from easy to more complex conditions'', which facilitates gradual establishment of few-shot capability in target domains that exhibit large gaps from the source domain~\cite{wang2021survey}. We propose Hybrid Progressive Augmentation (HPA) to generate more complex query views as the adaptation proceeds and the model becomes more capable, as illustrated in Fig.~\ref{fig:framework}~(yellow box). Specifically, we increase the task complexity through two designs:

\noindent \textbf{Augment more challenging views.} As the adaptation proceeds, we progressively apply stronger augmentation operations. At the beginning, we apply simple augmentations such as flipping to generate the first query image $I_{q_1}$. Subsequently, increasingly complex transformations are introduced to generate subsequent query images, ensuring that augmentations across views become progressively more complex through cumulative operations. For example, $I_{q_1}$ uses only flipping, $I_{q_2}$ further adds brightness adjustment, ..., and $I_{q_6}$ combines all previous operations and grid shuffle together (refer to Fig.~\ref{fig:framework}~(yellow box)).

\noindent \textbf{Increase the number of views.} Augmenting additional views and establishing support-query correspondence among all of them implicitly increases the challenge, as prediction errors accumulate and propagate to subsequent views (refer to Tab.~\ref{tab:dmp_seq}). $N$ is set to 1 at the beginning, only requiring the model to predict accurate mask of $I_{q_1}$ based on $I_s$. As the adaptation proceeds, $N$ is adaptively increased, allowing for the augmentation of more query images. Subsequently, the model is expected to accurately predict across all $N$ augmented query images $\{I_{q_i}\}_{i=1}^N$.

\subsection{Dual-chain Multi-view Prediction}\label{ssec:dmp}

Even though we obtain $N$ query images, establishing few-shot capability remains challenging, as an effective strategy for leveraging these augmented views is essential. Consequently, we design a dual-chain strategy to comprehensively exploit these views: a sequential multi-view prediction chain and a parallel one (refer to Fig.~\ref{fig:framework}~(red box)).

\noindent \textbf{Sequential multi-view predictions and supervisions.} It is straightforward to propose a chain that includes $I_s$ and all \(\{I_{q_i}\}_{i=1}^N\), requiring the model to predict accurate masks across all of them sequentially. Based on the success of \cite{nie2024cross,fan2024adapting}, we incorporate SSP~\cite{fan2022self} to enhance prediction accuracy. Support prototype $P_s$ and prototype of the first query $P_{q_1}^{seq}$ are derived from support feature $F_s$, support mask $M_s$, and query feature $F_{q_1}$\footnote{We use the superscript ``seq'' (abbr. for sequential) to indicate that the variables or results are obtained within the sequential prediction chain. Similarly, the superscript ``par'' (abbr. for parallel) signifies that variables or results are from the parallel chain.}:
\vspace{-1mm}
\begin{equation}\label{eqn:ps}
    P_s = MAP(F_s, M_s),\ \ \ 
    P_{q_1}^{seq} = SSP(F_{q_1}, P_s),
\end{equation}
where $MAP$ is masked average pooling operation. More details of SSP are in the supplementary materials. 

Next, we predict support mask $\hat{M}_s$ and mask $\hat{M}_{q_1}^{seq}$ for the first query image:
\vspace{-1mm}
\begin{equation}\label{eqn:predicted_ms}
    \hat{M}_s = \sigma(\cos(F_s, P_s)),\ \ \ 
    \hat{M}_{q_1}^{seq} = \sigma(\cos(F_{q_1}, P_{q_1}^{seq})),
\end{equation}
where $\cos$ is cosine similarity, and $\sigma$ represents softmax function. Concurrently, support base loss $\mathcal{L}_{bs}$ and sequential query loss $\mathcal{L}_{q_1}^{seq}$ for the first query image are computed:
\vspace{-1mm}
\begin{equation}\label{eqn:base_loss}
    \mathcal{L}_{bs} = \mathcal{L}_{bce}(\hat{M}_s, M_s),\ \ \ 
    \mathcal{L}_{q_1}^{seq} = \mathcal{L}_{bce}(\hat{M}_{q_1}^{seq}, M_{q_1}),
\end{equation}
where $\mathcal{L}_{bce}$ is the binary cross entropy loss. Drawing inspiration from the effectiveness of bi-directional predictions in alleviating overfitting~\cite{zhang2021few,nie2024cross}, we also incorporate reverse prediction, where the roles of the support and query images are exchanged, with the query image guiding the segmentation of the support:
\vspace{-1mm}
\begin{equation}\label{eqn:seq_s1}
    \begin{aligned}
        P_s^{{seq}_1} = SSP(F_s, P_{q_1}^{seq}),& \ 
        \hat{M}_s^{{seq}_1} = \sigma(Sim(F_s, P_s^{{seq}_1})),\\
        \mathcal{L}_s^{{seq}_1} =\ &\mathcal{L}_{bce}(\hat{M}_s^{{seq}_1}, M_{s}),\\
    \end{aligned}
\end{equation}
where $P_s^{{seq}_1}$ is another support prototype predicted from $I_{q_1}$\footnote{A pseudo prototype of the support image, \(P_s^{{seq}_1}\), can be predicted from the predicted pseudo prototype \(P_{q_1}^{seq}\) of the first query image. \(P_s^{{seq}_1}\) differs from \(P_s\), which is obtained from the ground-truth support mask \(M_s\). We include \(i\) in the superscript to indicate that this support prototype is derived from the predicted result of the \(i^{th}\) query image.}. $\mathcal{L}_s^{{seq}_1}$ provides additional regularization, aiding in establishing few-shot capability with limited data. With additional query images introduced by HPA, the above predictions can be seamlessly extended into a sequential chain. 

Specifically, for the $j^{th}\ (1\leq j\leq N)$ query image $I_{q_{j}}$, we obtain $\mathcal{L}_{q_{j}}^{seq}$, and $\mathcal{L}_s^{{seq}_{j}}$ from $P_{s}^{{seq}_{(j-1)}}$($P_{s}^{{seq}_{0}} = P_s$ in Eq.~\ref{eqn:ps}), with details in the supplementary materials. The results from Eqs.~\ref{eqn:ps}, \ref{eqn:predicted_ms} and \ref{eqn:seq_s1} correspond to the case when $j=1$. Ideally, a well-trained model should make accurate predictions on all queries sequentially~\cite{nie2024cross}. Otherwise, errors in earlier predictions propagate, affecting subsequent results, while the corresponding loss provides additional regularization for target domains.

\noindent \textbf{Parallel multi-view predictions and supervisions.} To directly mimic ``support-to-query'' predictions during inference~\cite{lei2022cross,herzog2024adapt,nie2024cross} and further establish correspondences between the accessible support image $I_s$ and query images \(\{I_{q_i}\}_{i=1}^N\), we propose a parallel multi-view prediction chain that operates with the sequential chain simultaneously. Specifically, it leverages $I_s$ to segment \(\{I_{q_i}\}_{i=1}^N\) in parallel ($1\leq i\leq N$) and includes the reverse predictions. Then, ground-truth masks provide supervision signals:
\vspace{-1mm}
\begin{equation}\label{eqn:par_pred}
    \begin{aligned}
        P_{q_i}^{par} = SSP(F_{q_i}, P_s),\ \ \ \ & \ \hat{M}_{q_i}^{par} = \sigma(\cos(F_{q_i}, P_{q_i}^{par})),\\
        P_s^{{par}_i} = SSP(F_s, P_{q_i}^{par}),& \ \hat{M}_s^{{par}_i} = \sigma(\cos(F_s, P_s^{{par}_i})),\\
        \mathcal{L}_{q_i}^{par} = \mathcal{L}_{bce}(\hat{M}_{q_i}^{par}, M_{q_i}&),\ \mathcal{L}_s^{{par}_i} = \mathcal{L}_{bce}(\hat{M}_s^{{par}_i}, M_s).\\
    \end{aligned}
\end{equation}

\subsection{Loss Function}\label{ssec:loss}
\textbf{Sequential loss.} We would like to clarify that $\mathcal{L}_{q_1}^{seq}$ and $\mathcal{L}_s^{{seq}_1}$ are identical to $\mathcal{L}_{q_1}^{par}$ and $\mathcal{L}_s^{{par}_1}$, respectively. Therefore, our proposed sequential loss starts from index $2$. The supervision from the sequential prediction chain can be summarized as follows:
\vspace{-1mm}
\begin{equation}\label{eqn:seq_loss}
    \mathcal{L}^{seq}\ = \sum_{i=2}^{\textit{N}}(\mathcal{L}_{q_i}^{seq}+\mathcal{L}_s^{{seq}_i}).
\end{equation}

\noindent \textbf{Parallel loss.} Similarly, the parallel supervision consists of two parts, parallel support loss and parallel query loss:
\vspace{-1mm}
\begin{equation}\label{eqn:par_loss}
    \mathcal{L}_s^{par}\ = \sum_{i=1}^{\textit{N}}\mathcal{L}_s^{{par}_i},\ \ \ 
    \mathcal{L}_q^{par}\ = \sum_{i=1}^{\textit{N}}\mathcal{L}_{q_i}^{par}.
\end{equation}

\noindent \textbf{Total loss.} The total loss combines and balances the previous loss terms $\mathcal{L}_{bs}$, $\mathcal{L}^{seq}$, $\mathcal{L}_s^{par}$, and $\mathcal{L}_q^{par}$:
\vspace{-1mm}
\begin{equation}\label{eqn:total_loss}
    \begin{aligned}
        \mathcal{L}\ = &\ \lambda_{bs}\times\mathcal{L}_{bs}+\lambda^{seq}\times\mathcal{L}^{seq} \\
         + &\ \lambda_s^{par}\times\mathcal{L}_s^{par}+\lambda_q^{par}\times\mathcal{L}_q^{par},
    \end{aligned}
\end{equation}
where $\lambda_{bs}=0.2$, $\lambda^{seq}=0.1$, $\lambda_s^{par}=0.4$, and $\lambda_q^{par}=1$. The determination of the values of these parameters is discussed in the supplementary materials. The overall MPA algorithm is presented step by step in the supplementary materials.

\subsection{Extension to $K$-shot Setting}\label{ssec:kshot}
In the extension to the $K$-shot (\(K > 1\)) setting, $K$ support images with their corresponding masks, denoted as \(S = \{(I_s^i, M_s^i)\}_{i=1}^K\), are provided for adaptation. All augmented queries \(\{(I_{q_i}, M_{q_i})\}_{i=1}^N\) are derived from a randomly selected support image-mask pair \(\{(I_s^i, M_s^i)\}\). MPA can be efficiently extended to this setting as follows. As described in Sec.~\ref{ssec:dmp}, the initial prediction utilizes the averaged support prototype \(\bar{P}_s = \frac{1}{K} \sum_{i=1}^{K} P_{s}^{i}\) to predict \(\hat{M}_q\). In the reversed prediction procedure, we use pseudo prototype \(P_{q_i}\) of $i^{th}$ query image to predict each \(\hat{M}_s^i\) in parallel.

\begin{table*}[t]
    \renewcommand\arraystretch{1.1}
    \centering
    \vspace{-4mm}
    \caption{Quantitative comparisons between the proposed MPA and existing methods over four widely adopted data-scarce domains (in mIoU (\%)). The best mIoU number under each setup is highlighted by \textbf{bold} font.}
    \vspace{-2mm}
    \resizebox{\linewidth}{!}{
    \setlength{\tabcolsep}{7pt}
    \begin{tabular}{c|l|cc|cc|cc|cc|cc}
        \toprule[1pt]
        \multirow{2}*{Backbone}& \multirow{2}*{Method}& \multicolumn{2}{c|}{Deepglobe}& \multicolumn{2}{c|}{ISIC}& \multicolumn{2}{c|}{Chest X-Ray}& \multicolumn{2}{c|}{FSS-1000}& \multicolumn{2}{c}{\textbf{Average}}\\\cline{3-12}
        ~& ~& 1-shot& 5-shot& 1-shot& 5-shot& 1-shot& 5-shot& 1-shot& 5-shot& 1-shot& 5-shot\\\hline
        \multirow{2}*{Vgg-16}& AMP~\cite{siam2019amp}& 37.6& 40.6& 28.4& 30.4& 51.2& 53.0& 57.2& 59.2& 43.6& 45.8\\
        ~& PATNet~\cite{lei2022cross}& 28.7& 34.8& 33.1& 45.8& 57.8& 60.6& 71.6& 76.2& 47.8& 54.4\\\hline
        \multirow{15}*{Res-50}& PGNet~\cite{zhang2019pyramid}& 10.7& 12.4& 21.9& 21.3& 34.0& 28.0& 62.4& 62.7& 32.2& 31.1\\
        ~& PANet~\cite{wang2019panet}& 36.6& 45.3& 25.3& 34.0& 57.8& 69.3& 69.2& 71.7& 47.2 & 55.1\\
        ~& CaNet~\cite{zhang2019canet}& 22.3& 23.1& 25.2& 28.2& 28.4& 28.6& 70.7& 72.0& 36.6& 38.0\\
        ~& RPMMs~\cite{yang2020prototype}& 13.0& 13.5& 18.0& 20.0& 30.1& 30.8& 65.1& 67.1& 31.6& 32.9\\
        ~& PFENet~\cite{tian2020prior}& 16.9& 18.0& 23.5& 23.8& 27.2& 27.6& 70.9& 70.5& 34.6& 35.0\\
        ~& RePRI~\cite{boudiaf2021few}& 25.0& 27.4& 23.3& 26.2& 65.1& 65.5& 71.0& 74.2& 46.1& 48.3\\
        ~& HSNet~\cite{min2021hypercorrelation}& 29.7& 35.1& 31.2& 35.1& 51.9& 54.4& 77.5& 81.0& 47.6& 51.4\\
        ~& PATNet~\cite{lei2022cross}& 37.9& 43.0& 41.2& 53.6& 66.6& 70.2& 78.6& 81.2& 56.1& 62.0\\
        ~& SSP~\cite{fan2022self}& 41.3& 54.2& 48.6& 65.4& 72.6& 73.0& 77.0& 79.4& 60.0& 68.0\\
        ~& DR-Adapter~\cite{su2024domain}& 41.3& 50.1& 40.8& 48.9& 82.4& 82.3& 79.1& 80.4& 60.9& 65.4\\
        ~& IFA~\cite{nie2024cross}& 50.6& 58.8& 66.3& 69.8& 74.0& 74.6& 80.1& \textbf{82.4}& 67.8& 71.4\\

        
        ~& ABCDFSS~\cite{herzog2024adapt}& 42.6& 49.0& 45.7& 53.3& 79.8& 81.4& 74.6& 76.2& 60.7& 65.0\\ 
        
        ~& \cellcolor{mycolor}\textbf{MPA (w/ Source-Training)}& 
        \cellcolor{mycolor}\textbf{54.2}& \cellcolor{mycolor}\textbf{60.8}& \cellcolor{mycolor}\textbf{74.3}& \cellcolor{mycolor}\textbf{74.4}& \cellcolor{mycolor}\textbf{89.1}& \cellcolor{mycolor}\textbf{91.0}& \cellcolor{mycolor}\textbf{81.4}& \cellcolor{mycolor}81.4& \cellcolor{mycolor}\textbf{74.8}& \cellcolor{mycolor}\textbf{76.9}\\
        
        ~& \cellcolor{mycolor}\textbf{MPA (w/o Source-Training)}& 
        \cellcolor{mycolor}\textbf{53.1}& \cellcolor{mycolor}\textbf{59.4}& \cellcolor{mycolor}\textbf{71.1}& \cellcolor{mycolor}\textbf{71.3}& \cellcolor{mycolor}\textbf{89.0}& \cellcolor{mycolor}\textbf{90.6}& \cellcolor{mycolor}\textbf{80.2}& \cellcolor{mycolor}80.8& \cellcolor{mycolor}\textbf{73.4}& \cellcolor{mycolor}\textbf{75.5}\\
        \bottomrule[1pt]
    \end{tabular}
    }
    \vspace{-4mm}
    \label{tab:cdfss}
\end{table*}

\begin{table}[t]
    \renewcommand\arraystretch{1.1}
    \centering
    \vspace{0mm}
    \caption{Quantitative comparisons between the proposed MPA and existing methods over SUIM (in mIoU (\%)). The best mIoU number under each setup is highlighted by \textbf{bold} font.}
    \vspace{-2mm}
    \resizebox{\linewidth}{!}{
    \setlength{\tabcolsep}{16.4pt}
    \begin{tabular}{l|c|c}
        \toprule[1pt]
        Method& 1-shot& 5-shot\\\hline
        HSNet~\cite{min2021hypercorrelation}& 28.8& -\\
        SCL~\cite{zhang2021self}& 31.8& -\\
        HDMNet~\cite{peng2023hierarchical}& 23.4& 30.9\\
        RemDiff~\cite{wang2022remember}& 34.7& -\\
        RestNet~\cite{huang2023restnet}& 25.2& -\\       
        PATNet~\cite{lei2022cross}& 32.1& 40.2\\
        PxMtch~\cite{chen2024pixel}& 34.8& -\\
        ABCDFSS~\cite{herzog2024adapt}& 35.1& 41.3\\
        \cellcolor{mycolor}\textbf{MPA (w Source-Training)}& \cellcolor{mycolor}\textbf{55.5}& \cellcolor{mycolor}\textbf{62.0}\\
        \cellcolor{mycolor}\textbf{MPA (w/o Source-Training)}& \cellcolor{mycolor}\textbf{54.2}& \cellcolor{mycolor}\textbf{61.1}\\
        \bottomrule[1pt]
    \end{tabular}
    }
    \vspace{-5mm}
    \label{tab:suim}
\end{table}
\section{Experiment}
\label{sec:exp}

\subsection{Datasets}
We conduct extensive experiments over five data-scarce target datasets, covering satellite images~\cite{demir2018deepglobe}, medical screenings~\cite{codella2019skin,tschandl2018ham10000,candemir2013lung,jaeger2013automatic}, tiny objects~\cite{li2020fss}, and underwater scenes~\cite{islam2020semantic}.
Fig.~\ref{fig:visualization} shows sample images from these datasets.
\textbf{Deepglobe}~\cite{demir2018deepglobe} is a satellite image dataset containing 6 terrain categories, such as urban, agriculture, rangeland, forest, water, and barren. For CD-FSS, we follow PATNet~\cite{lei2022cross} to split images into smaller pieces. \textbf{ISIC2018}~\cite{codella2019skin,tschandl2018ham10000} comprises medical images of skin lesions. It captures three types of skin lesions. \textbf{Chest X-Ray}~\cite{candemir2013lung, jaeger2013automatic} is collected for Tuberculosis test. Its grayscale images enhance the diversity of evaluation. \textbf{FSS-1000}~\cite{li2020fss} comprises natural images of everyday objects. It poses significant challenges due to its scarce samples and tiny objects in images. \textbf{SUIM}~\cite{islam2020semantic} consists of 7 categories of underwater objects, including fish, plants, divers, robots, ruins, and rocks.

\subsection{Implementation Details}
\label{ssec:exp_detail}
We adopt the ResNet-50~\cite{he2016deep} pre-trained on ImageNet~\cite{deng2009imagenet} as the backbone model. Consistent with previous works~\cite{fan2022self,nie2024cross}, we discard the last stage and last ReLU for better generalization. The model is implemented with PyTorch~\cite{paszke2017automatic}. Following PATNet~\cite{lei2022cross}, we resize all images to $400\times400$ to reduce memory consumption and accelerate the learning. The learning rate is set as 5e-4 for all datasets. 
For data augmentation in Sec.~\ref{ssec:hca}, we adopt a set of transformations in PyTorch~\cite{paszke2017automatic}, including horizontal-flip, vertical-flip, 90-degree rotation, brightness-variation, and hue-variation. We adopt an adaptive criterion to dynamically increase the task complexity during progressive adaptation. Specifically, when the performance stagnates for three consecutive epochs, we regard it as performance saturation and introduce an additional, more challenging query view. This new view cumulatively incorporates all previously used augmentation operations along with an additional, more complex operation. The mIoU~\cite{min2021hypercorrelation} serves as the evaluation metric.

\subsection{Comparison with State-of-the-art Methods}
\begin{figure*}[t]
    \centering
    \vspace{-6mm}
    \includegraphics[width=\linewidth]{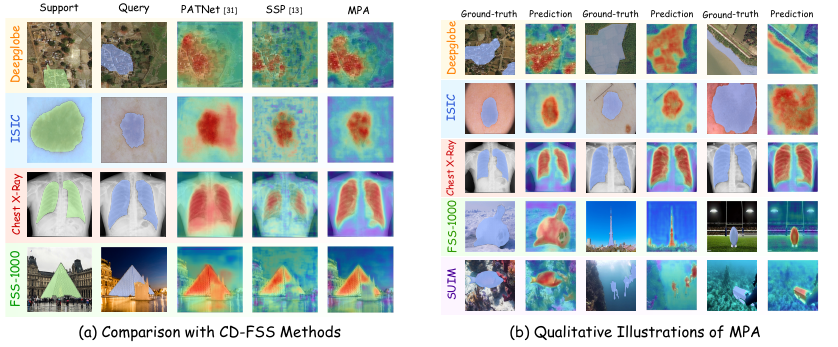}
    \vspace{-7mm}
    \caption{Qualitative illustrations over five data-scarce domains, including Deepglobe, ISIC, Chest X-Ray, FSS-1000, and SUIM from up to down. Segmentation comparisons with state-of-the-art methods are in (a). For each pair of support image (ground truth highlighted by green) and query image (ground truth highlighted by blue), Column 3-5 show the corresponding segmentation heatmaps by PATNet~\cite{lei2022cross}, SSP~\cite{fan2022self}, and our MPA. Segmentation heatmaps of MPA over samples from five domains are shown in (b). Best viewed in color.}
    \vspace{-2mm}
    \label{fig:visualization}
\end{figure*}

\textbf{Effectiveness.} 
Tabs.~\ref{tab:cdfss} and \ref{tab:suim} compare MPA with the state-of-the-art methods, showing that MPA consistently achieves superior performance by a substantial margin. In particular, it surpasses IFA~\cite{nie2024cross} by 7.0\% (1-shot) and 5.5\% (5-shot) in mIoU. For the five target benchmarks, Deepglobe~\cite{he2016deep} is unique due to its aerial view and complex background. MPA can handle such a complex segmentation task well and improve the mIoU by 3.6\% under the 1-shot setup. The medical images in ISIC~\cite{codella2019skin,tschandl2018ham10000} and Chest X-Ray~\cite{candemir2013lung,jaeger2013automatic} have a clean background, and their target regions occupy a large portion of the image. MPA also segments such images well. FSS-1000~\cite{li2020fss} is more challenging due to substantial variations across images in hundreds of categories. Though existing methods benefit from a small domain gap with respect to the source domain (Pascal VOC~\cite{everingham2010pascal}), MPA still excels under the 1-shot setup. For the very different underwater images in SUIM~\cite{islam2020semantic}, MPA achieves new state-of-the-art performance as well. The extensive experiments over these different benchmarks demonstrate the superior robustness and generalization capability of the proposed MPA. As shown in Tab.~\ref{tab:sam}, MPA also outperforms SAM-based methods in terms of both effectiveness and model size, highlighting its overall superiority. In addition, we perform qualitative comparisons to verify the superiority of the MPA over multiple CD-FSS benchmarks. As Fig.~\ref{fig:visualization}(a) shows, MPA in the Column 5 achieves clearly better segmentation as compared with the baseline SSP~\cite{fan2022self} in the Column 4. Fig.~\ref{fig:visualization}(b) and Figure in the supplementary materials show more visualizations, further demonstrating the great segmentation results of the proposed MPA.

\begin{table}[t]
    \renewcommand\arraystretch{1.1}
    \centering
    \vspace{0mm}
    \caption{Quantitative comparisons between the proposed MPA and SAM-based methods (in mIoU (\%)). The best mIoU number under each setup is highlighted by \textbf{bold} font.}
    \vspace{-2mm}
    \resizebox{\linewidth}{!}{
    \setlength{\tabcolsep}{8.2pt}
    \begin{tabular}{l|c|c|c|c}
        \toprule[1pt]
        Method& Deepglobe& ISIC& FSS-1000& \textbf{Avg.}\\\hline
        APSeg~\cite{he2024apseg}& 35.9& 45.4& 79.7& 53.7\\
        TAVP~\cite{yang2024tavp}& 46.1&	54.9& 79.1&	60.0\\
        PerSAM~\cite{zhang2023personalize}& 31.4&	23.9&	71.2&	42.2\\
        \cellcolor{mycolor}\textbf{MPA (w/o Source)}& \cellcolor{mycolor}\textbf{53.1}& \cellcolor{mycolor}\textbf{71.1}& \cellcolor{mycolor}\textbf{80.2}& \cellcolor{mycolor}\textbf{68.1}\\
        \bottomrule[1pt]
    \end{tabular}
    }
    \vspace{-3mm}
    \label{tab:sam}
\end{table}

\noindent\textbf{Discussion.}
Previous methods typically rely on source-domain training to establish few-shot capability. However, as discussed earlier, MPA is designed to progressively establish few-shot capability with augmented query views in target domains. To further validate its effectiveness, we conduct experiments without source-training, directly adapting the backbone model to target domains with MPA. Remarkably, this single-stage adaptation achieves performance comparable to the conventional two-stage pipeline, attaining 73.4\% mIoU in the 1-shot setting and 75.5\% in the 5-shot setting on average. Compared with IFA~\cite{nie2024cross}, MPA without source training improves mIoU by 5.6\% (1-shot) and 4.1\% (5-shot), while delivering substantial average gains of 12.7\% (1-shot) and 10.5\% (5-shot) over the source-free ABCDFSS~\cite{herzog2024adapt}. These results confirm that most performance gains are derived from the adaptation stage, validating that designing suitable adaptation strategies with limited accessible data is the key to CD-FSS. In addition, removing the source-training stage reduces total training time by approximately 80\%, substantially improving efficiency. Consequently, all subsequent experiments are reported under the source-free setting by default, with a detailed discussion of the efficiency benefits in Sec.~\ref{ssec:analysis}.

\begin{table}[t]
    \renewcommand\arraystretch{1.1}
    \centering
    \vspace{-1mm}
    \caption{Ablation studies for technical designs of MPA. Our proposed HPA and DMP designs both show significant effectiveness.}
    \vspace{-2mm}
    \resizebox{\linewidth}{!}{
    \setlength{\tabcolsep}{21pt}
    \begin{tabular}{l|c|c}
        \toprule[1pt]
        Incorporated design& Deepglobe& ISIC\\\hline
        Baseline& 42.1& 42.2\\
        $+$ HPA& 47.8& 61.2\\
        \cellcolor{mycolor}\textbf{$+$ DMP $+$ HPA}& \cellcolor{mycolor}\textbf{53.1}& \cellcolor{mycolor}\textbf{71.1}\\
        \bottomrule[1pt]
    \end{tabular}
    }
    \label{tab:comp}
    \vspace{-1mm}
    
\end{table}

\subsection{Ablation Studies}
\label{ssec:ab-study}
\noindent\textbf{Technical designs ablation.} 
We conduct ablation experiments over our designed Hybrid Progressive Augmentation (HPA) and Dual-chain Multi-view Prediction (DMP) techniques. We use SSP~\cite{fan2022self} with the pre-trained ResNet-50~\cite{he2016deep} as the baseline. First, we incorporate HPA (Sec.~\ref{ssec:hca}) by progressively increasing the complexity of augmented views, while limiting the number of views during training. As shown in Tab.~\ref{tab:comp}, this yields performance gains of 5.7\% and 19.0\% on the two datasets, respectively. 
Then, we incorporate HPA and DMP, and observe additional performance gains of 5.3\% and 9.9\% on the two datasets. Notably, DMP serves as the cornerstone of the MPA framework, effectively building diverse prediction scenarios and achieving larger improvements by exhaustively exploiting the multiple views generated by HPA. These two designs complement each other to progressively establish few-shot capability. Overall, the results underscore the importance of augmenting more views and designing appropriate strategies for adaptation.

\begin{figure}[t]
    \centering
    \vspace{-4mm}
    \includegraphics[width=\linewidth]{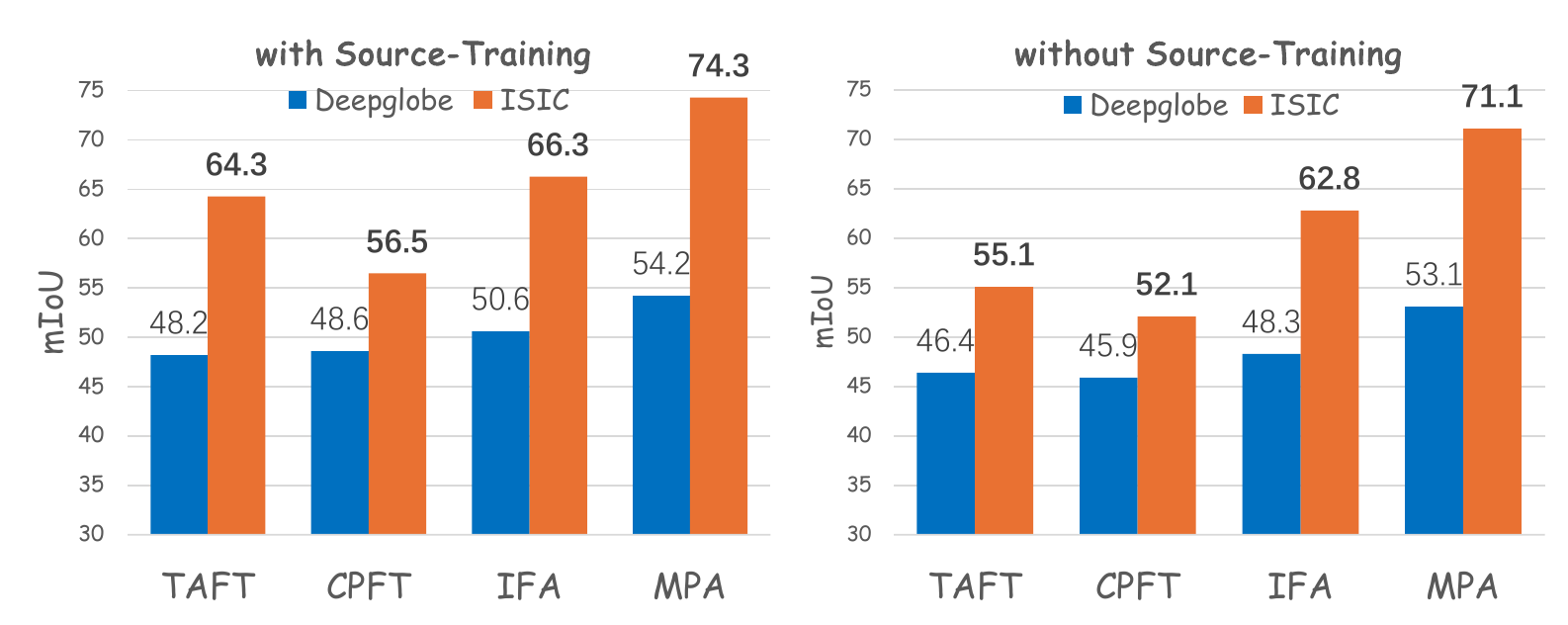}
    \vspace{-7mm}
    \caption{Ablation study on different adaptation strategies.}
    \vspace{-5mm}
    \label{fig:ab_ft}
\end{figure}

\noindent\textbf{Adaptation strategies.} 
We benchmark MPA against several widely adopted adaptation strategies from prior studies. For a fair comparison, all strategies are evaluated under both source-trained and direct adaptation setups. Specifically, Task-Adaptive Fine-Tuning (TAFT)~\cite{lei2022cross} works at the meta-testing stage, which leverages Kullback-Leibler divergence loss to minimize the distance between the category prototypes of the segmented query image and the support set. Copy-Paste Fine-Tuning (CPFT)~\cite{gao2022acrofod} mixes the foreground region of a target image with the background region of a random source image as the support set, forcing source-to-target adaptation as inspired by \cite{ghiasi2021simple}. IFA~\cite{nie2024cross} establishes support-query correspondences during the fine-tuning stage. It augments a single view of the given support image only and is susceptible to overfitting. As Fig.~\ref{fig:ab_ft} shows, MPA delivers consistently strong performance and remains competitive across both setups. These results highlight that MPA does not depend heavily on source-domain training and can reliably establish few-shot capability in the target domain using only limited accessible data.

\begin{table}[t]
    \renewcommand\arraystretch{1.1}
    \centering
    \vspace{-4mm}
    \caption{Effectiveness of progressive strategies. Both explicit (generating more challenging query views) and implicit (increasing the number of augmented views) progressive augmentation benefit the final performance.}
    \vspace{-2mm}
    \resizebox{\linewidth}{!}{
    \setlength{\tabcolsep}{29pt}
    \begin{tabular}{l|c}
        \toprule[1pt]
        Setups& mIoU \\\hline
        always 1 augmented query& 50.5\\
        \cellcolor{mycolor}\textbf{implicit progressive strategy}& \cellcolor{mycolor}\textbf{52.0}\\\hline
        always simple augmentation& 51.3\\
        \cellcolor{mycolor}\textbf{explicit progressive strategy}& \cellcolor{mycolor}\textbf{52.4}\\\hline
        \cellcolor{mycolor}\textbf{combine both strategies}& \cellcolor{mycolor}\textbf{53.1}\\
        \bottomrule[1pt]
    \end{tabular}
    }
    \vspace{-4mm}
    \label{tab:curr}
\end{table}

\noindent\textbf{Progressive strategies.} We examine the effectiveness of explicit and implicit progressive strategies as in Tab.~\ref{tab:curr}. The implicit progressive strategy increases the number of augmented queries as performance improves, which inherently accumulates prediction errors and leads to inaccuracy in the subsequent queries~\cite{nie2024cross} (in the sequential chain of DMP). Nevertheless, such inaccuracy facilitates the evaluation of the generalization capability~\cite{nie2024cross}, and more discussion about this implicit design is the supplementary materials. As Tab.~\ref{tab:curr} shows, the implicit progressive improves the mIoU by around 1.4\%. The explicit progressive strategy generates more challenging queries progressively via stronger augmentations, \textit{e.g.}, from simple flipping to complex color jittering. As Tab.~\ref{tab:curr} shows, the explicit progressive strategy brings in around 1.1\% improvement in mIoU. More detail of the involved hyper-parameters is discussed in the supplementary materials.

\subsection{Analysis}
\label{ssec:analysis}

\noindent\textbf{Analysis of DMP.} We perform an in-depth analysis of the DMP design. For the parallel chain, the augmented query views create multiple independent learning routes, effectively increasing the data available for mimicking the ``support-to-query'' prediction. This leads to clear performance gains by increasing the amount of adaptation data. For the sequential chain, errors can accumulate and propagate across multiple prediction turns. As training progresses, later views undergo stronger augmentations, resulting in increasingly perturbed predictions that deviate from the ground truth (see Tab.~\ref{tab:dmp_seq}). To tackle these errors, we apply dense supervision to every sequentially augmented prediction, encouraging the model to learn representations robust to a broad spectrum of perturbations. This design improves the model's generalization potential. Qualitative results in the supplementary material further support the motivation of the sequential prediction chain.

\noindent\textbf{Analysis of HPA.} We conduct adaptation under three setups: (1) using only simple augmentation, (2) replacing simple augmentation with a single, more complex augmentation per view, and (3) our augmentation accumulation strategy. The results in Tab.~\ref{tab:hca_aug_ft} show that our adopted augmentation accumulation strategy achieves the best results.

\begin{table}[t]
    \renewcommand\arraystretch{1.1}
    \centering
    \vspace{-4mm}
    \caption{The mIoU (\%) under fine-tuning with different augmentation strategies. Our proposed cumulative augmentation strategy achieves the best performance.}
    \vspace{-2mm}
    \resizebox{\linewidth}{!}{
    \setlength{\tabcolsep}{16pt}
    \begin{tabular}{l|c|c}
        \toprule[1pt]
        Augmentation& Deepglobe& ISIC\\\hline
        Simple augmentation& 51.3& 67.9\\
        Augmentation replacement& 51.9& 68.5\\
        \cellcolor{mycolor}\textbf{Ours}& \cellcolor{mycolor}\textbf{53.1}\ \ & \cellcolor{mycolor}\textbf{71.1}\\
        \bottomrule[1pt]
    \end{tabular}
    }
    \label{tab:hca_aug_ft}
    \vspace{0mm}

\end{table}

\noindent\textbf{Efficiency.}
We benchmark the training time of MPA (w/o source-training) with several representative methods. As Tab.~\ref{tab:efficiency} shows, MPA saves over 80\% time as compared with the state-of-the-art training-required method IFA but achieves clearly better segmentation results\footnote{MPA performs adaptation with one image under 1-shot setting, after which it is evaluated on the entire test set.}. Additionally, data augmentation in MPA is performed only once at the beginning, which introduces negligible overhead. Further, feature extraction of all augmented views in MPA runs in parallel without incurring extra computational cost, and all experiments can be performed on GTX 2080Ti GPUs.

\begin{table}[t]
    \renewcommand\arraystretch{1.1}
    \centering
    \vspace{-2mm}
    \caption{MPA significantly reduces the training time (min) and improves the mIoU performance (\%).}
    \vspace{-2mm}
    \resizebox{\linewidth}{!}{
    \setlength{\tabcolsep}{11pt}
    \begin{tabular}{l|cc|cc}
        \toprule[1pt]
        \multicolumn{5}{c}{\textbf{\textit{Adaptation with Source-Training}}}\\\hline
        \multirow{2}*{Adaptation strategy}& \multicolumn{2}{c|}{Deepglobe}& \multicolumn{2}{c}{ISIC}\\\cline{2-5}
        ~& time& mIoU &time& mIoU\\\hline
        PATNet~\cite{lei2022cross}& 538 & 37.9& 532 &41.2\\
        IFA~\cite{nie2024cross}& 555& 50.6& 551& 66.3\\\hline
        \multicolumn{5}{c}{\textbf{\textit{Adaptation without Source-Training}}}\\\hline
        \cellcolor{mycolor}\textbf{MPA}& \cellcolor{mycolor}\textbf{98}& \cellcolor{mycolor}\textbf{53.1}& \cellcolor{mycolor}\textbf{95}& \cellcolor{mycolor}\textbf{71.1}\\
        \bottomrule[1pt]
    \end{tabular}
    }
    \label{tab:efficiency}
    \vspace{-3mm}
    
\end{table}

\section{Conclusion}
In this paper, we tackle the challenges of Cross-Domain Few-Shot Segmentation, where large domain gaps and scarce target-domain data hinder the establishment of few-shot capability. We proposed Multi-View Progressive Adaptation (MPA), which gradually builds few-shot capability from both the data and strategy perspectives. By introducing Hybrid Progressive Augmentation to gradually generate more complex and diverse views, and Dual-chain Multi-view Prediction to exhaustively exploit them, MPA effectively establish few-shot capability in target domains. Extensive experiments demonstrate its superiority across multiple benchmarks. In future work, we plan to extend the progressive adaptation framework to broader cross-domain tasks and explore its potential in real-world applications with more complex domain shifts.

\section*{Acknowledgement}
This study is supported by the Collaborative Initiative, Interdisciplinary Graduate Programme, Nanyang Technological University, and the Ministry of Education Singapore, under the Tier-2 project scheme with a project number MOE-T2EP20123-0003.
{
    \small
    \bibliographystyle{ieeenat_fullname}
    \bibliography{main}
}


\end{document}